\documentclass[conference]{IEEEtran}

\usepackage{graphicx}
\usepackage{booktabs}
\usepackage{amsmath,amssymb}
\usepackage{url}
\usepackage{tikz}
\usetikzlibrary{arrows.meta,positioning,fit,backgrounds,calc}
\usepackage[hidelinks]{hyperref}
\usepackage{cite}

\newcommand{\code}[1]{\texttt{\small #1}}

\title{Closed-Loop Physics-Based Manipulation Evaluation at a Commodity
Price: What 4.4 L4 GPU hours at \$3.50 Buys You}
\author{\IEEEauthorblockN{Rohun Kshirsagar\IEEEauthorrefmark{1} and Diana Rypkema\IEEEauthorrefmark{1}}
\IEEEauthorblockA{IROS 2026 Scale-Infra Workshop\\
\IEEEauthorrefmark{1}Equal contribution}}

\begin{document}
\maketitle

\begin{abstract}
\small
Running evaluations of contact-rich robotic manipulation at statistical scale
has largely been possible only at well-resourced labs, highlighting the need for
affordable and efficient options for broader audiences. We present a reproducible
evaluation harness wrapper based on a previous study at 200-rollout scale on
serverless commodity GPUs, at a cost of approximately \$3.50~USD and 38 minutes
of wall-clock per replication (4.4 L4-GPU-hours across eight parallel shards).
Using the harness, we report a partially controlled negative result on action chunking with
sparse data: three independently trained chunked checkpoints produce zero
genuine successes across 780 rollouts, while three single-step checkpoints
range from 2.6\% to 11.8\% success rate. A per-step-replanning control partially isolates chunk prediction, rather than open-loop chunk execution as the mechanism, with the failure localized to the training formulation and multi-step prediction. This work further democratizes physics-based policy development and evaluation through our reusable harness.
\end{abstract}

\section{Introduction}

Progress of faithful physics-based
simulation~\cite{lbm,robomimic,simpler,drake,sap,mujoco,zong} with
affordable and efficient GPU access has created a unique opportunity for a
broader audience outside of top robotics academic and industry labs to tackle
hard problems in contact-rich robotic manipulation. Access to GPUs has
historically been a bottleneck~\cite{computedivide}, but services like Modal
have recently made it affordable and easy to access a cluster of GPUs without
needing to work with quotas, networking, and IAM permissions on major cloud
providers like Azure, Amazon Web Services, and Google Cloud Platform.
Additionally, simulation benchmarks in this space are increasingly distributed
as Python packages (e.g.~\cite{lbm}), which enjoy a large number of users
compared to packages written in other languages. The reproducibility literature
in machine learning identifies environment and dependency reconstruction as a
first-order technical barrier~\cite{semmelrock}, but the concrete work it
entails is rarely documented. In our case, it meant GPU-container configuration,
headless rendering setup, and policy-serving plumbing.

For this study, we stood up an evaluation harness wrapper that runs the benchmark
from Barreiros et al.~\cite{lbm} end-to-end on serverless commodity GPUs,
trained our own policies for a single simulation task, and ran repeated
200-rollout closed-loop evaluations for \$3.49~USD each, at 38 minutes of
wall-clock time. (The benchmark's own documentation notes that a CPU-based
OpenGL stack yields very slow evaluation throughput, which makes GPU access a
practical prerequisite for this workload rather than an optimization.) Thanks to
affordable and efficient GPU access, our constraints were not compute but
undocumented environment requirements --- an EGL vendor ICD file, an interpreter
path assumption in an asset download script, and checkpoint formats that drifted
between training and serving. Our contributions are: (1) a released serverless
evaluation harness wrapper and reporting protocol for 200-rollout closed-loop
evaluation of a simulated task at commodity cost and (2) a controlled experiment
showing that action chunking degrades policy performance when the training data
is sparse.

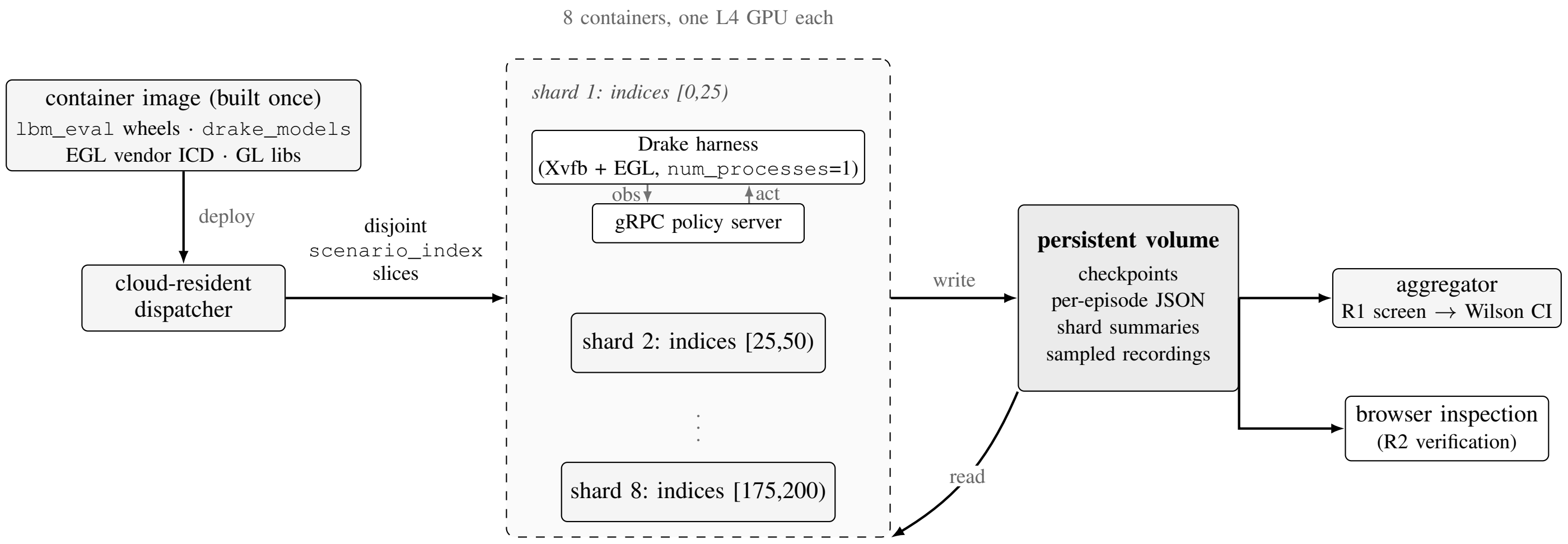
\begin{figure*}[t]
\centering
\begin{tikzpicture}[
  font=\footnotesize,
  box/.style={draw, rounded corners=2pt, align=center, minimum height=6mm,
              inner sep=3pt},
  shard/.style={box, fill=black!4, minimum width=30mm},
  inner/.style={draw, rounded corners=1.5pt, fill=white, align=center,
                font=\scriptsize, minimum height=4.6mm, inner sep=2pt,
                minimum width=25mm},
  store/.style={box, fill=black!8, minimum width=26mm, minimum height=22mm},
  flow/.style={-{Latex[length=1.6mm]}, thick},
  faint/.style={-{Latex[length=1.4mm]}, semithick, black!55},
  lbl/.style={font=\scriptsize, black!60, align=center},
]

\node[inner] (drake1) {Drake harness\\\scriptsize(Xvfb + EGL, \texttt{num\_processes}=1)};
\node[inner, below=2.2mm of drake1] (srv1) {gRPC policy server};
\begin{scope}[on background layer]
  \node[shard, fit=(drake1)(srv1), inner sep=3.2pt] (s1) {};
\end{scope}
\node[font=\scriptsize\itshape, black!70, above=0.8mm of s1.north west,
      anchor=south west] (s1lab) {shard 1: indices [0,25)};
\draw[faint, transform canvas={xshift=-6mm}] (drake1.south) -- (srv1.north)
  node[midway, left, font=\scriptsize, black!60, xshift=0.5mm] {obs};
\draw[faint, transform canvas={xshift=6mm}] (srv1.north) -- (drake1.south)
  node[midway, right, font=\scriptsize, black!60, xshift=-0.5mm] {act};

\node[shard, below=7mm of s1, minimum height=7mm] (s2)
  {shard 2: indices [25,50)};
\node[lbl, below=1.6mm of s2] (dots) {$\vdots$};
\node[shard, below=1.2mm of dots, minimum height=7mm] (s8)
  {shard 8: indices [175,200)};

\begin{scope}[on background layer]
  \node[draw, dashed, rounded corners=3pt, fill=black!2,
        fit=(s1lab)(s1)(s2)(s8), inner sep=5pt] (fanout) {};
\end{scope}
\node[lbl, above=2.5mm of fanout.north] {8 containers, one L4 GPU each};

\node[box, left=26mm of fanout, minimum width=24mm, fill=black!4] (disp)
  {cloud-resident\\dispatcher};
\draw[flow] (disp) -- node[midway, above=0.8mm, align=center,
  font=\scriptsize, text=black]
  {disjoint\\\texttt{scenario\_index}\\slices} (fanout);

\node[store, right=15mm of fanout] (vol)
  {\textbf{persistent volume}\\[2pt]
   \scriptsize checkpoints\\
   \scriptsize per-episode JSON\\
   \scriptsize shard summaries\\
   \scriptsize sampled recordings};
\draw[flow] (fanout) -- node[lbl, above, midway] {write} (vol);
\draw[flow] (vol.south west) to[bend left=18]
  node[lbl, midway, fill=white, inner sep=1pt] {read} (fanout.south east);

\node[box, right=11mm of vol, minimum width=24mm, fill=black!4] (agg)
  {aggregator\\\scriptsize R1 screen $\rightarrow$ Wilson CI};
\node[box, below=8mm of agg, minimum width=24mm] (insp)
  {browser inspection\\\scriptsize (R2 verification)};
\draw[flow] (vol) -- (agg);
\draw[flow] (vol.east) |- (insp.west);

\node[box, above=11mm of disp, minimum width=30mm, fill=black!4, align=center]
  (img) {container image (built once)\\[1pt]
         \scriptsize \texttt{lbm\_eval} wheels $\cdot$ \texttt{drake\_models}\\
         \scriptsize EGL vendor ICD $\cdot$ GL libs};
\draw[flow] (img.south) -- (disp.north)
  node[lbl, midway, right, xshift=0.5mm] {deploy};

\end{tikzpicture}
\caption{Runtime architecture of one 200-rollout replication. A
cloud-resident dispatcher fans out eight containers, each evaluating a
disjoint slice of \texttt{scenario\_index}; because the harness derives each
rollout's random seed from its index, the union of shards reproduces the
unsharded run exactly. Within a container the Drake harness renders headless
under Xvfb and queries a co-located gRPC policy server, which keeps the policy
implementation outside the harness process. Each shard writes per-episode
results to a persistent volume; screening and aggregation read them back
afterwards, so no result depends on the dispatching process surviving the run.
Model assets and the EGL vendor ICD are baked into the image at build time.}
\label{fig:arch}
\end{figure*}

\section{Methods}

\subsection{Fan-Out Architecture}

\begin{figure}[t]
\centering
\includegraphics[width=\columnwidth]{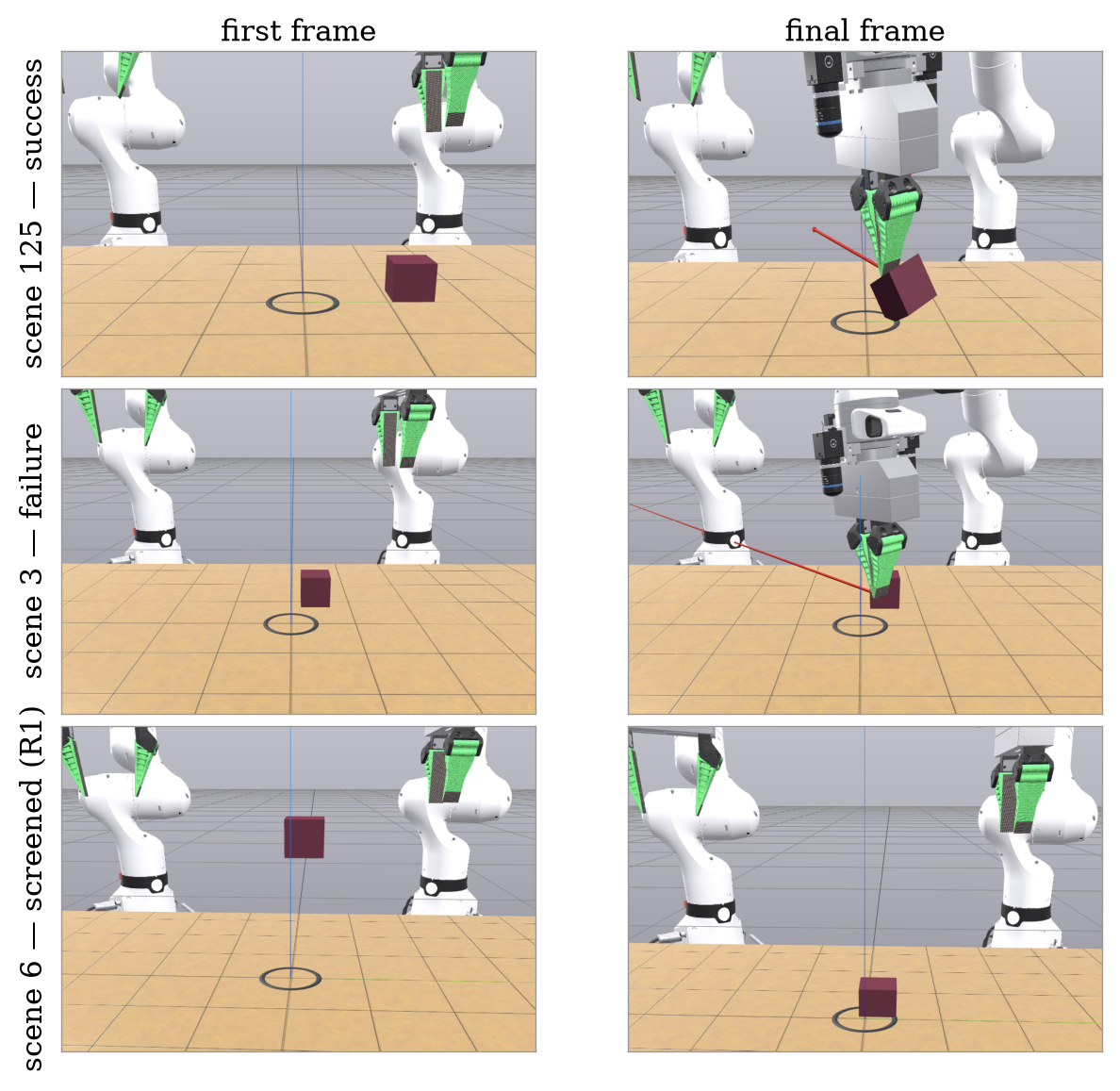}
\caption{First and final frames of three rollouts under the v1 baseline are shown in the columns for three different scenes (rows). In each scene, the policy instructs the right robotic arm to pick up the red box and place it in the interior of the circular goal marker on the tabletop and the rollout is judged as a success or failure. In
scene 125 the arm transports the box onto the goal marker and the rollout is
scored successful. In scene 3 the arm reaches toward the box, overshoots, and
holds position for the remainder of the episode without relocating it. Red
arrows are Drake hydroelastic contact-force vectors. Scene 6 is one of five
scenarios whose initial condition places the box directly above the goal; it
registers success within 0.3~s with both arms stationary, and is excluded under
R1. Camera position varies slightly between recordings.}
\label{fig:rollouts}
\end{figure}

We chose to base our evaluation harness wrapper off of Barreiros et
al.~\cite{lbm} because it is among the most statistically rigorous evaluations
of simulated and real-world large behavior models to date, providing a realistic
physics-based benchmark that requires GPUs for efficiently training and
evaluating policies. Barreiros et al.~\cite{lbm} used two Franka FR3 robot arms
in a Drake tabletop bimanual manipulation simulation. We evaluated on
\code{pick\_and\_place\_box}, a skill shipped with the \code{lbm\_eval} benchmark
GitHub repository and recommended by its README as a starting point~\cite{lbm}.
The task \code{pick\_and\_place\_box} evaluated the right Franka arm successfully
picking up a randomly placed box on the table and placing it into a circular
region in the middle of the table within 15 seconds (see
Fig.~\ref{fig:rollouts}). It did not appear in the paper's simulation evaluation
suite (Barreiros et al.~\cite{lbm} Table~S2), and the paper reports no success
rate for it, making it well-suited to infrastructure and ablation work.

Our evaluation runs as a serverless fan-out to containers on Modal
(Fig.~\ref{fig:arch}). Drake's model assets are baked into the container image
at build time, and rendering runs headless under Xvfb with EGL/OpenGL. A
replication is sharded by disjoint \code{scenario index} slices, one shard per
container; because the harness derives each rollout's random seed from its
index, the union of shards reproduces the unsharded run exactly. Within a shard,
evaluations run serially, so parallelism comes from shard count alone. The
policy is served over gRPC by a standalone process that the harness connects to
as a client, which keeps the policy outside the harness and makes one checkpoint
servable under different execution parameters without retraining. Each shard
writes per-episode results to a persistent volume; aggregation reads them back
afterwards, so no result depends on the dispatching process surviving the run.
Two build-time steps were required and were not documented anywhere: the GPU
runtime
mounts the NVIDIA driver but not the userspace GL libraries, so the image writes
the EGL vendor ICD file Drake's renderer expects; and Drake's asset-download
script assumes an interpreter path the base image does not use, so the build
adds a verified symlink. The stack is otherwise a Debian-slim base, Python 3.12,
the four public \code{lbm\_eval} wheels, and PyTorch.

\subsection{Reporting Protocol}

The success rates of the single-task policies we trained were low due to having
only 100 samples from the TRI repository \code{lbm\_eval}~\cite{lbm}. To account for evaluation artifacts that
could swing the result, we adopted three rules:

\textbf{R1:} completions recorded in under one second are excluded as benchmark
artifacts. Five of the 200 scenarios (indices 6, 94, 116, 140, 146) succeed at
approximately 0.3 seconds before the policy meaningfully acts, while all failures continue through to the 15 second limit, validating the one second rule.
The same five, and only those
five, appear in every evaluation we retain results for, across both decoder
families and multiple independently trained checkpoints, indicating a property
of the benchmark's initial conditions rather than of any policy. All results are reported over the resulting
195-scene denominator.

\textbf{R2:} every claimed success is verified by visually inspecting its recording,
never on the harness flag alone.

\textbf{R3:} results are reported as $N/195$ with a 95\% Wilson score
confidence interval~\cite{wilson}, preferred to the normal approximation because
it remains well-behaved at proportions near zero. The full per-episode JSON is
retained and released. Omitting R1 would have reported a 2.5\% success rate from artifacts alone for a
policy that does nothing.

\subsection{Policy Descriptions}

We note that all models share a common encoder: an ImageNet-pretrained ResNet-18
on a single camera view, concatenated with a Multi-Layer Perceptron (MLP)
embedding of the 18-dimensional end-effector pose, predicting a 20-dimensional
action (bimanual end-effector poses in position and 6D rotation, plus two
gripper widths)~\cite{robomimic,dp}. They are trained from the same preprocessed
data (99 usable episodes, 4,203 keyframes) and differ only in decoder family and
in the pair $(H, K)$, where $H$ is the number of actions predicted per inference
and $K$ the number executed before replanning: policy ``v1'' is a Behavior
Cloning (BC) regressor with Mean Squared Error (MSE) loss at $H{=}1$; policy ``v2a'' a single-step
Denoising Diffusion Probabilistic Model (DDPM), which was also used in Barreiros
et al.~\cite{lbm}; policy v2b a chunked DDPM at $(16, 8)$; and policy v2b-bc is
a chunked BC regressor at $H{=}8$, sharing v1's decoder family with targets
sliced from v2b's windows. Seeded rows are independently trained checkpoints
differing only in initialization and batch order; the train/validation split is
fixed across all rows. For reference, per-task demonstration counts in the
source suite range from 49 to 490, with most tasks at 196.

\section{Experiments and Results}

\subsection{Commodity Cost and Wall-Clock Time}

\begin{table}[t]
\centering\small
\caption{Cost of one 200-rollout closed-loop replication, from per-shard
\code{elapsed\_time}. The final row is the compute required to reproduce every
trained variant and every replication reported in Table~\ref{tab:ladder} from
the released code, assuming clean execution.}
\label{tab:cost}
\begin{tabular}{@{}ll@{}}
\toprule
Wall-clock (fan-out) & 37.7 min (8 parallel shards) \\
Eval compute & 4.36 L4-GPU-hr \\
Cost (USD) & \$3.49 at \$0.80 / L4-hr \\
Mean rollout duration & $\approx$78 s \\
Train + one replication & $\approx$\$3.85 \\
\midrule
All work in Table~\ref{tab:ladder} & $\approx$\$34 to reproduce \\
\bottomrule
\end{tabular}
\end{table}

As stated earlier, improving the success rate of \code{pick\_and\_place\_box} was
not the goal of this project, but rather the intention was to measure the cost
and speed to implement and test using the evaluation harness wrapper.
Table~\ref{tab:cost} reports representative cost and runtime of v1 (first BC
model), highlighting the opportunity to train quickly and inexpensively using our
evaluation harness wrapper. A single training + 200 rollout evaluation costs
approximately \$3.85~USD, of which \$3.49~USD is evaluation. Training takes 26
minutes while evaluation takes 38 minutes total wall-clock time when sharded out
to 8 L4 GPUs. The entire study, including environment bring-up, every trained
variant, and every replication in Table~\ref{tab:ladder}, cost approximately
\$34~USD.

\begin{table}[!b]
\centering\small
\caption{Ablation ladder on \code{pick\_and\_place\_box}, 195 non-trivial
scenes, 95\% Wilson confidence intervals. Every chunked checkpoint scored zero across 780
rollouts; every single-step checkpoint scored above zero. Validation MSE does
not predict closed-loop success: 0.0148 and 0.0152 correspond to 11.8\% and
2.6\%. Success rates for all 200 runs, including the five trivial success are in (). $\dagger$~DDPM underfits (sampled MSE above the noise baseline).}
\label{tab:ladder}
\begin{tabular}{@{}llcrl@{}}
\toprule
Checkpoint & Decoder & $(H,K)$ & Success Rate & 95\% Wilson CI \\
\midrule
v1 (orig.)   & BC   & (1,1)  & 5.1\% (7.5\%)  & [2.8, 9.2] \\
v1 seed1     & BC   & (1,1)  & 11.8\% (14\%) & [8.0, 17.1] \\
v1 seed2     & BC   & (1,1)  & 2.6\%  (5\%) & [1.1, 5.9] \\
\midrule
v2a          & DDPM & (1,1)  & 0\% (2.5\%)   & [0, 1.9] \\
v2b$^\dagger$& DDPM & (16,8) & 0\% (2.5\%)   & [0, 1.9] \\
\midrule
v2b-bc       & BC   & (8,4)  & 0\% (2.5\%)   & [0, 1.9] \\
v2b-bc       & BC   & (8,1)  & 0\% (2.5\%)   & [0, 1.9] \\
v2b-bc seed1 & BC   & (8,4)  & 0\% (2.5\%)   & [0, 1.9] \\
v2b-bc seed2 & BC   & (8,4)  & 0\% (2.5\%)   & [0, 1.9] \\
\bottomrule
\end{tabular}
\end{table}

\subsection{Negative Result in the Sparse Data Regime}

Table~\ref{tab:ladder} suggests that multi-step prediction and the training formulation, including action chunking, contributes to policy performance
collapse: no chunked checkpoint produced a genuine completion across four
evaluations and 780 rollouts, while all three single-step checkpoints scored
above zero. We report this at the checkpoint level rather than pooling rollouts,
since rollouts within a checkpoint are not independent. The served chunk-trained
policy (v2b-bc) at $K{=}1$ shows a 0\% success rate while all three v1
checkpoints have a non-zero performance, indicating action chunking contributes to performance degradation. We note that validation MSE does not
translate to success rate (2.6\% vs 11.8\% success rate for similar MSE), and
chunked checkpoints all receive 0\%. Also, using rule R2 (visual inspection)
shows that the baseline fails by compounding drift, while the chunked
checkpoints fail by generating trajectories that do not track the scene even
when replanned every step.

\section{Conclusion and Future Directions}

We presented a reproducible serverless evaluation harness wrapper that runs a
Drake-based closed-loop manipulation benchmark at 200-rollout scale from outside
the originating lab for under \$3.50~USD and forty minutes per replication.
Compute was not the obstacle, thanks to affordably available serverless GPUs like
Modal; undocumented environment requirements and format drift between training
and serving were the limitations. Running the evaluation carefully changed what
would have been reported: 5 of 200 scenarios succeeded before the policy acted,
and a standard design choice, action chunking, contributed to policy performance collapse across three independently trained checkpoints, with a
per-step-replanning control partially locating the damage in chunk prediction. The study
covered one task, one dataset scale, and one model-size regime, and the chunking
result should not be assumed to generalize beyond it. Action chunking is well-supported in data-rich settings~\cite{dp,act}, but with roughly one hundred
demonstrations and a policy succeeding a few percent of the time, it may hurt the
policy during chunk prediction rather than in execution. Future studies should expand the harness to include other single and multi-task benchmarks. The harness would benefit from bringing in new data for training better policies, including using compliant-contact simulation like in Castro et al.~\cite{sap} to
generate synthetic training data to create policies with a
higher success rate on tasks like \code{pick\_and\_place\_box}, but instead applied to multi-task learning.

Our work will enable a broader audience outside of major robotics labs to accurately evaluate their own policies in service of new science and building new applications and products not currently considered by the market.  We believe empowering them will make for rapid iterations that would otherwise be slow (or infeasible) for them to perform without our harness.

\section*{Code and Data Availability}
The code from this paper can be found in this repo on GitHub:
\url{https://github.com/rohungry/evaluation_harness_wrapper}, including the Modal image definition specifying the container environment and package versions, the training scripts and configs, trained checkpoints, and per-episode evaluation outputs for every reported run. The BC and DDPM policies evaluated here were trained by us on demonstration data publicly available from TRI; the benchmark software and data are TRI's, and no artifact is withheld for licensing reasons. The repository documents the full environment specification and reproduction steps.

\section*{Disclosure}

We proposed ideas, wrote and edited the code, ran the code and inspected
artifacts on our local machine and on the server, evaluated the results by manual
inspection, and wrote the outline and prose of the manuscript. An LLM coding
assistant was used for drafting code, infrastructure scaffolding, debugging, and drafting this
manuscript. All reported results were produced by the released code and verified
by the probes and evaluation protocol described in Sections 2 through 4.

\section*{Acknowledgment}
We thank Vatshank Chaturvedi for helpful conversations on the harness design and scaling experiments, and three anonymous reviewers for their thoughtful comments.

\bibliographystyle{IEEEtran}
\bibliography{refs}

@article{lbm,
  author  = {Barreiros, Jose and Beaulieu, Andrew and Bhat, Aditya and
             Cory, Rick and Cousineau, Eric and Dai, Hongkai and
             Fang, Ching-Hsin and Hashimoto, Kunimatsu and
             Irshad, Muhammad Zubair and Itkina, Masha and others},
  title   = {A Careful Examination of Large Behavior Models for Multitask
             Dexterous Manipulation},
  journal = {Science Robotics},
  volume  = {11},
  number  = {113},
  pages   = {eaea6201},
  year    = {2026}
}

@inproceedings{robomimic,
  author    = {Mandlekar, Ajay and Xu, Danfei and Wong, Josiah and
               Nasiriany, Soroush and Wang, Chen and Kulkarni, Rohun and
               Fei-Fei, Li and Savarese, Silvio and Zhu, Yuke and
               Mart{\'i}n-Mart{\'i}n, Roberto},
  title     = {What Matters in Learning from Offline Human Demonstrations for
               Robot Manipulation},
  booktitle = {Proc. Conference on Robot Learning (CoRL)},
  year      = {2021}
}

@inproceedings{simpler,
  author    = {Li, Xuanlin and Hsu, Kyle and Gu, Jiayuan and Pertsch, Karl and
               Mees, Oier and Walke, Homer Rich and Fu, Chuyuan and
               Lunawat, Ishikaa and Sieh, Isabel and Kirmani, Sean and others},
  title     = {Evaluating Real-World Robot Manipulation Policies in Simulation},
  booktitle = {Proc. Conference on Robot Learning (CoRL)},
  year      = {2024}
}

@article{dp,
  author  = {Chi, Cheng and Xu, Zhenjia and Feng, Siyuan and
             Cousineau, Eric and Du, Yilun and Burchfiel, Benjamin and
             Tedrake, Russ and Song, Shuran},
  title   = {Diffusion Policy: Visuomotor Policy Learning via Action
             Diffusion},
  journal = {The International Journal of Robotics Research},
  volume  = {44},
  number  = {10-11},
  pages   = {1684--1704},
  year    = {2025}
}

@article{act,
  author  = {Zhao, Tony Z. and Kumar, Vikash and Levine, Sergey and
             Finn, Chelsea},
  title   = {Learning Fine-Grained Bimanual Manipulation with Low-Cost
             Hardware},
  journal = {arXiv preprint arXiv:2304.13705},
  year    = {2023}
}

@electronic{drake,
  author       = {Tedrake, Russ and {the Drake Development Team}},
  title        = {Drake: Model-Based Design and Verification for Robotics},
  howpublished = {Software},
  year         = {2019},
  url          = {https://drake.mit.edu}
}

@article{wilson,
  author  = {Wilson, Edwin B.},
  title   = {Probable Inference, the Law of Succession, and Statistical
             Inference},
  journal = {Journal of the American Statistical Association},
  volume  = {22},
  number  = {158},
  pages   = {209--212},
  year    = {1927}
}

@article{sap,
  author  = {Castro, Alejandro M. and Permenter, Frank N. and Han, Xuchen},
  title   = {An Unconstrained Convex Formulation of Compliant Contact},
  journal = {IEEE Transactions on Robotics},
  volume  = {39},
  number  = {2},
  pages   = {1301--1320},
  year    = {2022}
}

@inproceedings{mujoco,
  author    = {Todorov, Emanuel and Erez, Tom and Tassa, Yuval},
  title     = {{MuJoCo}: A Physics Engine for Model-Based Control},
  booktitle = {Proc. IEEE/RSJ International Conference on Intelligent Robots
               and Systems (IROS)},
  pages     = {5026--5033},
  year      = {2012}
}

@inproceedings{zong,
  author    = {Zong, Zeshun and Jiang, Chenfanfu and Han, Xuchen},
  title     = {A Convex Formulation of Frictional Contact for the Material
               Point Method and Rigid Bodies},
  booktitle = {Proc. IEEE/RSJ International Conference on Intelligent Robots
               and Systems (IROS)},
  pages     = {1831--1838},
  year      = {2024}
}

@article{computedivide,
  author  = {Ahmed, Nur and Wahed, Muntasir},
  title   = {The De-democratization of {AI}: Deep Learning and the Compute
             Divide in Artificial Intelligence Research},
  journal = {arXiv preprint arXiv:2010.15581},
  year    = {2020}
}

@article{semmelrock,
  author  = {Semmelrock, Harald and Ross-Hellauer, Tony and Kopeinik, Simone
             and Theiler, Dieter and Haberl, Armin and Thalmann, Stefan and
             Kowald, Dominik},
  title   = {Reproducibility in Machine-Learning-Based Research: Overview,
             Barriers, and Drivers},
  journal = {AI Magazine},
  volume  = {46},
  number  = {2},
  pages   = {e70002},
  year    = {2025}
}

\end{document}